\documentclass[a4paper,twoside]{article}
\usepackage[top=3.3cm, bottom=4.2cm, left=2.6cm, right=2.6cm]{geometry}
\usepackage{epsfig}
\usepackage{subcaption}
\usepackage{calc}
\usepackage{amssymb}
\usepackage{amstext}
\usepackage{amsmath}
\usepackage{amsthm}
\usepackage{multicol}
\usepackage{pslatex}
\usepackage{apalike}
\usepackage[bottom]{footmisc}
\usepackage{multirow}
\usepackage{xcolor}
\usepackage[colorlinks=true, allcolors=blue]{hyperref}

\usepackage{SCITEPRESS}

\begin{document}

\title{Performance Evaluation and Comparison of a New Regression Algorithm}

\author{\authorname{Sabina Gooljar, Kris Manohar and Patrick Hosein}
\affiliation{Department of Computer Science, The University of the West Indies, St. Augustine, Trinidad}
\email{sabinagooljar@gmail.com, Kris.Manohar@sta.uwi.edu, Patrick.Hosein@sta.uwi.edu}
}

\abstract{In recent years, Machine Learning algorithms, in particular supervised learning techniques, have been shown to be very effective in solving regression problems. We compare the performance of a newly proposed regression algorithm against four conventional machine learning algorithms namely, Decision Trees, Random Forest, $k$-Nearest Neighbours and XG Boost. The proposed algorithm was presented in detail in a previous paper but detailed comparisons were not included. We do an in-depth comparison, using the Mean Absolute Error (MAE) as the performance metric, on a diverse set of datasets to illustrate the great potential and robustness of the proposed approach. The reader is free to replicate our results since we have provided the source code in a GitHub repository while the datasets are publicly available.}

\onecolumn \maketitle \normalsize \setcounter{footnote}{0} \vfill

\section{\uppercase{Introduction}}
\label{sec:introduction}

Machine Learning algorithms are regularly used to solve a plethora of regression problems. The demand for these algorithms has increased significantly due to the push of digitalisation, automation and analytics. Traditional techniques such as Random Forest, Decision Trees and XG Boost have been integral in various fields such as banking, finance, healthcare and engineering. Technology is always evolving and technological advancements are driven by factors such as human curiosity, problem-solving and the desire for increased efficacy and reliability. Researchers are constantly working on improving these existing methods as well as exploring new improved strategies as can be seen in \cite{Hosein2022}. This approach uses a distance metric (Euclidean distance) and a weighted average of the target values of all training data points to predict the target value of a test sample. The weight is inversely proportional to the distance between the test point and the training point, raised to the power of a parameter $\kappa$. In our paper, we investigate the performance of this novel approach and several well-established machine learning algorithms namely XG Boost, Random Forest, Decision Tree and $k$-NN using the Mean Absolute Error (MAE) as the performance metric. We intend to showcase the potential of this new algorithm to solve complex regression tasks across diverse datsets. In the next section, we describe related work and then the theory of the proposed approach. After, we present and discuss the findings such as any issues encountered. Finally, we advocate that the proposed approach may be robust and efficient making it extremely beneficial to the field.
 
\section{\uppercase{Related Work and Contributions}}
\label{sec:related work and contribution}

In this section, we summarize the various regression techniques we considered and then discuss differences with the proposed approach. Our contribution, which is a detailed comparison, is then outlined.

\subsection{Decision Tree}
Decision Tree is a supervised machine learning algorithm that uses a set of rules to make decisions. Decision Tree algorithm starts at the root node where it evaluates the input features and selects the best feature to split the data \cite{Quinlan1986}. The data is split in such a way so that it minimises some metric that quantifies the difference between the actual values and the predicted values such as Mean Squared Error and Sum of Squared error. Then a feature and a threshold value is chosen that best divides the data into two groups. The data is split recursively into two subsets until a stopping condition is met such as having too few samples in a node. When the decision tree is constructed, predictions are made by traversing from the root node to a leaf node. The predicted value is calculated as the mean of the target values in the training samples which is associated with that leaf node.

\subsection{Random Forest}
Random Forest builds decision trees on different samples and then averages the outputs for regression tasks \cite{Breiman2001}. It works on the principle of an ensemble method called Bagging. Ensemble is combining multiple models and then the set of models is used to make predictions instead of using an individual model. Bagging which is also known as Bootstrap Aggregation selects random samples from the original dataset. Each model is created from the samples that are given by the original data with replacement. Individual decision trees are constructed for each sample and each tree then produces its own output. These outputs are numerical values. The final output is then calculated from the average of these values which is known as aggregation. 

\subsection{ \textit{k}-Nearest Neighbours}
\textit{k}-Nearest Neighbours (\textit{k}NN) is a supervised machine learning algorithm that is used to solve both classification and regression tasks. Firstly, choose the number of neighbours (\textit{k}) which is used when making predictions. Then it calculates the distances (Euclidean) between a new query and the existing data and selects the specified number of neighbours (\textit{k}) that is closest to the query and finds the average of these values. The average is the predicted value.

\subsection{XGBoost}
XGBoost (eXtreme Gradient Boosting) is an ensemble learning algorithm that combines the output of weak learners (usually decision trees) to make more accurate predictions \cite{Chen2016}. New weak learners are added to the model iteratively with each tree aiming to correct the errors made by the previous learners. The training process is stopped when there is no significant improvement in a predefined number of iterations.

\subsection{New Algorithm}
Regression models enable decision-making in a wide range of applications such as finance, healthcare, education and engineering. It is imperative that these regression models are precise and robust so that better decisions can be made to enhance and improve these fields. While there are various popular machine learning algorithms for solving regression tasks, we introduce a new regression model that shows high accuracy and robustness, ensuring that real-world applications are optimised. The core of the approach is similar to $k$-NN but instead of using samples in a neighbourhood, all samples are used and closer samples are weighted more heavily than those further away. In this case, there is no parameter $k$ to specify but we do introduce a parameter $\kappa$ that dictates the rate of decay of the weighting. 

\section{\uppercase{Proposed Approach}}
\label{sec: proposed}

The proposed approach was originally designed to determine a suitable insurance policy premium \cite{Hosein2022}. Specifically, \cite{Hosein2022} noted as personlisation increases (i.e., more features), predictions become less robust due to the reduction in the number of samples per feature, especially in smaller datasets. His main goal was to achieve an optimal balance between personlisation and robustness. Instead of using the samples available for each feature, his algorithm computes the weighted average of the target variable using all samples in the dataset. This algorithm uses the Euclidean distance metric and a hyperparameter $\kappa$, which controls the influence of the distance (the weights) between points in the data. Another aspect is that the same unit of distance is used for each feature which allows one, for example, to compare distance between a gender feature with the distance between an age feature. 

The $\kappa$ parameter introduced in \cite{Hosein2022} is used as an exponent in the weighting formula, where the weights are inversely proportional to the distance between data points raised to the power of kappa. When $\kappa$ is large, the influence of points further away from the test point decreases quickly since their distance raised to a large power becomes very large which in turn makes the weight very small. However, when $\kappa$ is small, the influence of points further away decreases slowly since the distance raised to a small power results in a relatively smaller value which then makes the inverse weight larger. 

The algorithm firstly normalises the ordinal features. Then the prediction is done in two parts. For example, say we have a single categorical feature \textit{Gender} with two values, \textit{Male} and \textit{Female}. In the first stage, we compute the mean for each feature value over all the training samples. That is, the average target value for all females \(\mu_{Gender,Female} \) and the average target value for all males \(\mu_{Gender,male} \). With these means, the distance \textit{d} between a \textit{Female} and a \textit{Male} is the distance between \(\mu_{Gender,Female} \) and \(\mu_{Gender,male} \). The second stage computes the prediction for a test sample. For a given test sample \textit{i}, its prediction is the weighted average of the target value over all training samples. These weights are computed as \( \frac{1}{(1 + d[i, j])^ {\kappa_{2}} }\), where  \( d[i, j] \) is the Euclidean distance between the test sample \textit{i} and the training sample \textit{j}.

 Numerical features pose an interesting challenge because they can have a wider (potentially infinite) range of values. For example, consider the feature of \textit{Age}. Compared to \textit{Gender}, this feature can easily span 40 values (e.g., 18 to 58) instead of two. Thus, for the same training set, the number of samples per age will be low which implies the means (i.e., \(\mu_{Age,20}\), \(\mu_{Age,21}\), \(\mu_{Age,30} \) etc.,) used to compute distances  (and hence the weights) may not be robust. Additionally, we may encounter some unique values in the testing data set that are not in the training data set or vice versa. Generally, categorical variables do not encounter these issues because the number of samples per category value is sufficient. In order to solve this problem for numerical features, we impute a value for means for each unique value in both the training and testing data sets.

These imputed means are calculated based on the distances between the attribute value and all training samples in the feature space. This distance assists the algorithm in determining which training samples are most relevant to the test point. Then the target values for these training samples are combined, using a weighted average similar to before. The weights are computed as \( \frac{1}{(1 + d_{f}[i, j])^{\kappa_{1}}} \), where  \( d_{f}[i, j] \) is the distance between the numerical feature values in test sample \textit{i} and training sample \textit{j}. For example, say \textit{f} = \textit{Age}, and test sample \textit{i}'s age is 30 and training sample \textit{j}'s age is 40 then \(d_{Age}[i, j] = |30 - 40| = 10 \).

This modification means our proposed approach uses two kappa values: one for the pre-processing (i.e., \( \kappa_{1}\)) and one for predicting (i.e., \(\kappa_{2}\)). We determine the optimal combination of kappa values (\(\kappa_{1}, \kappa_{2} ) \) that minimises the error. According to \cite{Hosein2022}, as \( \kappa_{2} \) increases, the error decreases up to a certain point and then the error increases after this point. Therefore, an optimal \( \kappa_{2} \) can be found that minimizes the error.  

We define a range of values for both the pre-processing and predicting parts. The initial range is determined through a trial and error process. We observe the MAE and adjust these values if needed. However, note that while the initial range of kappa values involves some trial and error, the process of finding the optimal combination within the range is essentially a grid search which is systematic and data-driven and ensures that the model is robust. Since the algorithm uses all the training data points in its prediction, it will be robust for small data sets or where there are not enough samples per category of a feature. The Pseudo code in Figure \ref{fig:pseudo-code} summarizes the steps of the proposed algorithm.

\begin{figure*}
\begin{align*}
  & 1: \quad C \equiv \text{set of categorical features} \\
  & 2: \quad O \equiv \text{set of ordinal features} \\
  & 3: \quad X \equiv \text{set of training samples} \\
  & 4: \quad Y \equiv \text{set of testing samples} \\
  & 5: \quad \kappa_{1}, \kappa_{2} > 0 \quad \text{tuning parameters}\\
  & 6: \quad \text{for each } f \in O \text{ do} \\
  & 7: \quad \quad \text{for each sample } j \text{ in } X \text{ do}  \\
  & 8:  \quad \quad \quad x_{\text{train}, j, f} \leftarrow \frac{x_{\text{train}, j, f}}{\max(f) - \min(f)}  \quad \text{(normalize feature values in the train set)}\\
  & 9: \quad \quad \text{end for} \\
  & 10:\quad \quad \text{for each sample } i \text{ in } Y \text{ do}  \\
  & 11:  \quad \quad \quad x_{\text{test}, i, f} \leftarrow \frac{x_{\text{test}, i, f}}{\max(f) - \min(f)} \quad \text{(normalize feature values in the test set)} \\
  & 12: \quad \quad \text{end for} \\
  & 13: \quad \text{end for} \\
  & 14: \quad vf \equiv \text{set of categories for feature } f \in C \\
  & 15: \quad \text{for each } f \in C \text{ do} \\
  & 16: \quad \quad \text{for each } v \in vf \text{ do} \\
  & 17: \quad \quad \quad z \equiv \{x \in X | x_{f} = v\} \\
  & 18: \quad \quad \quad \mu_{f, v} \leftarrow \frac{1}{|z|} \sum_{x \in z} y_x \quad (\text{mean target value over training samples where feature } f \text{ has value } v) \\
  & 19: \quad \quad \text{end for} \\
  & 20: \quad \quad \text{Replace category values with their mean } \mu_{f, v} \text{ in both } X \text{ and } Y \\
  & 21: \quad \text{end for} \\
  & 22: \quad \text{for each} f \in O \text{ do} \\ 
  & 23: \quad \quad \text{for sample } i \text{ with unique feature value } v \text{ in } X \text{ and } Y \text{ do} \\
  & 24: \quad \quad \quad \mu_{f, v} \leftarrow \frac{\sum_{j \in X} \frac{y_j}{(1 + d_{f}[i, j])^{\kappa_{1}}}}{\sum_{j \in X} \frac{1}{(1 + d_{f}[i, j])^{\kappa_{1}}}} \quad (\text{imputed mean target value over training samples when feature } f \text{ has value } v) \\
  & 25: \quad \quad \text{end for} \\
  & 26: \quad \quad \text{Replace feature values with the imputed mean values } \mu_{f, v} \text{ in both } X \text{ and } Y \\
  & 27: \quad \text{end for} \\
  & 28: \quad \text{for each test sample } i \text{ in } Y \text{ do} \\
  & 29: \quad \quad \text{for each training sample } j \text{ in } X \text{ do} \\ 
  & 30: \quad \quad \quad d[i, j] \leftarrow \left(\sum_{f \in F} (x_{\text{test}, i, f} - x_{\text{train}, j, f})^2 \right)^{\frac{1}{2}} \quad \text{(calculate Euclidean distance) } \\
  & 31: \quad \quad \text{end for} \\
  & 32: \quad \text{end for} \\
  & 33: \quad \text{for each test sample } i \text{ in } Y \text{ do} \\
  & 34: \quad \quad c[i] \leftarrow \frac{\sum_{j \in X} \frac{y_j}{(1 + d[i, j])^{\kappa_{2}}}}{\sum_{j \in X} \frac{1}{(1 + d[i, j])^{\kappa_{2}}}} \\
  & 35: \quad \text{end for} \\
  & 36: \quad \text{Return the output values } c
\end{align*}
\caption{Pseudo code for the algorithm.}
\label{fig:pseudo-code}
\end{figure*}

\section{\uppercase{Numerical Results}}

In this section, we describe the datasets that were used
and apply the various techniques in order to compare their
performances. A GitHub repository, \cite{Gooljar2023} containing the code used
in this assessment has been created to facilitate replication and
validation of the results by readers.

\subsection{Data Set Description}

The data sets used were sourced from the University of California at Irvine (UCI) Machine Learning Repository. We used a wide variety of data sets to illustrate the robustness of our approach. We removed samples with any missing values and encoded the categorical variables. No further pre-processing was done so that the results can be easily replicated. \autoref{tab1data} shows a summary of the data sets used.

\begin{table*}[ht]
\setlength{\tabcolsep}{2pt}
\caption{Summary of Data sets.}\label{tab1data} \centering
\renewcommand{\arraystretch}{1.5}
\begin{tabular}{|l|c|c|l|l|}
  \hline
  Dataset & No. of Samples & No. of Attributes & Target Value & Citation \\
  \hline
  Student Performance & 394 & 32 & G3 & \cite{Cortez2014} \\
  \hline
  Auto & 392 & 6 & mpg & \cite{Quinlan1993} \\
  \hline
  Energy Y2 & 768 & 8 & Y2 & \cite{Tsanas2012} \\
  \hline
  Energy Y1 & 768 & 8 & Y1 & \cite{Tsanas2012} \\
  \hline
  Iris & 150 & 4 & Sepal Length & \cite{Fisher1988} \\
  \hline
  Concrete & 1030 & 8 & Concrete Compressive Strength & \cite{Yeh2007} \\
  \hline
  Wine Quality & 1599 & 11 & Residual Sugar & \cite{Cortez2009} \\
  \hline
\end{tabular}
\end{table*}

\subsection{Feature Selection}

There are various ways to perform feature selection \cite{Banerjee2020} but the best subset of features can only be found by exhaustive search. However, this method is computationally expensive so we select the optimal subset of features for the Random Forest model using Recursive Feature Elimination with Cross-Validation(RFECV) \cite{Brownlee2021} and use these features for all other models. Note that for each model, there may be a different optimal subset of features and, in particular, this subset of features may not be optimal for the proposed approach so it is not provided with any advantage. \autoref{tab2feature} shows the selected attributes for each dataset. The optimal subset of features was the full set of features for Auto, Energy Y2 and Iris datasets. The columns are indexed just as they appear in the datasets from UCI.

\begin{table*}[ht]
\setlength{\tabcolsep}{20pt}
\caption{Summary of Selected Features (categorical features are color coded in red)}\label{tab2feature}
\centering
\renewcommand{\arraystretch}{1.5}
\begin{tabular}{|c|c|}
  \hline
  Dataset & Selected Attributes \\
  \hline
  Student Performance & [2,23,28,29,31] \\
  \hline
  Auto & [1,2,3,4,5,6] \\
  \hline
  Energy Y2 & [0,1,2,3,4,5,6,7] \\
  \hline
  Energy Y1 & [0,1,2,3,4,6] \\
  \hline
  Iris & [1,2,3, \textcolor{red}{4}] \\
  \hline
  Concrete & [0,1,2,3,4,6] \\
  \hline
  Wine Quality & [0,4,5,6,9] \\
  \hline
\end{tabular}
\end{table*}

\subsection{Performance Results}

We show the performances of the different algorithms Random Forest, Decision Tree, \textit{k}-Nearest Neighbors, XG Boost, and the proposed method. The models are evaluated  on seven datasets (Auto, Student Performance, Energy Y2, Energy Y1, Iris, Concrete, and Wine Quality). We used Mean Absolute Error (MAE) to measure the performance since it is robust and easy to interpret. It measures the average magnitude of errors between the predicted values and the actual values. Lower MAE values indicate better performance. 
We used 10 Fold cross-validation for each model to ensure that we achieve a more reliable estimation of the model's performance. 

\begin{figure*}[ht]
  \includegraphics[width=\textwidth]{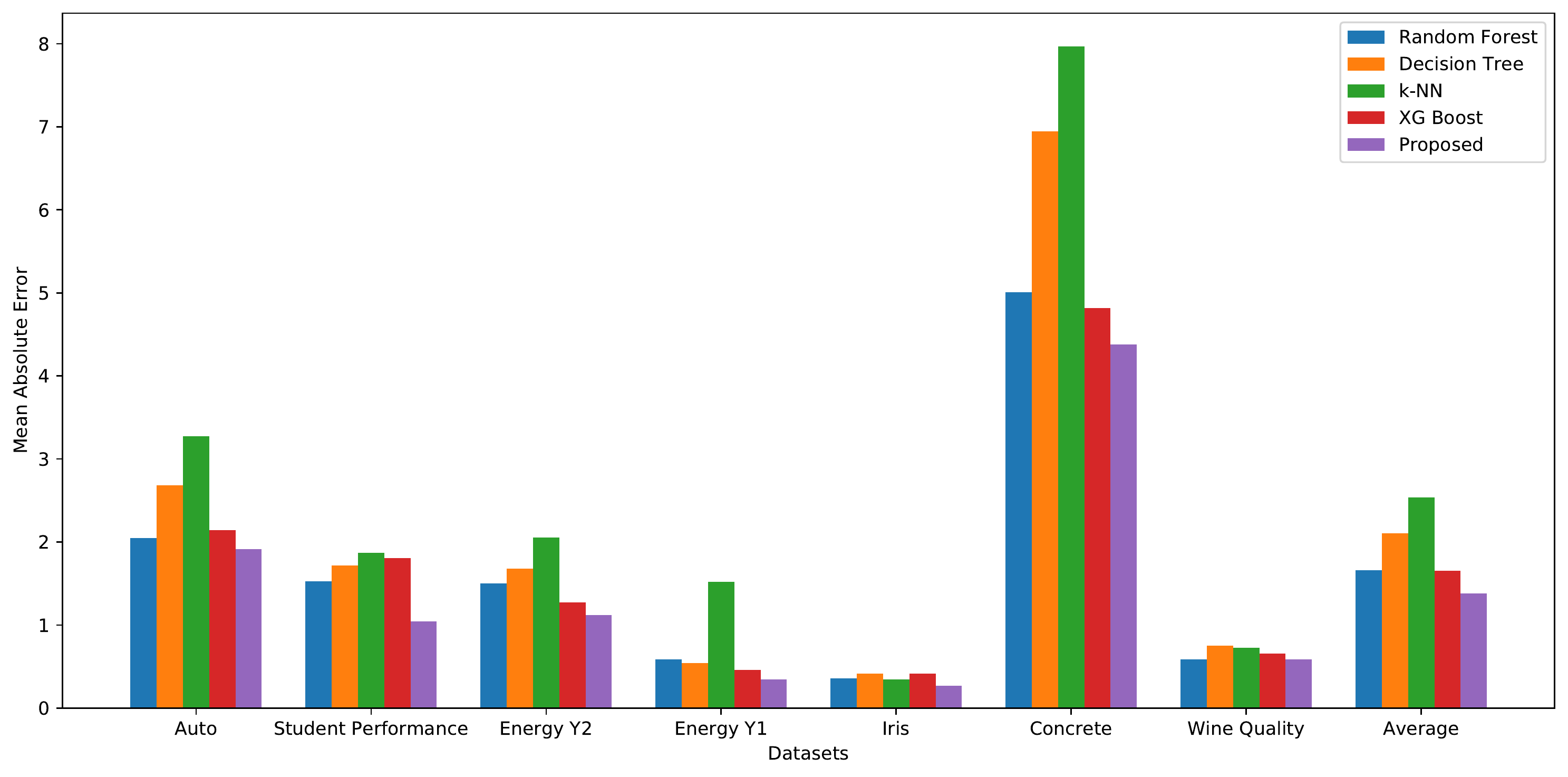}
  \caption{Mean Absolute Error for each Model across all Datasets.}
  \label{fig:datasets}
\end{figure*}

The performance results (MAE) for each algorithm with each dataset are provided in \autoref{tab:dataset_results}. The average MAE for the proposed algorithm is 1.380, while the next best performer is XG Boost, with an average MAE of 1.653. Random Forest, Decision Tree, and \textit{k}-NN have higher average MAEs of 1.659, 2.105 and 2.537, respectively. The proposed algorithm consistently performs better than the popular algorithms. We also provide a bar chart showing a comparison between the algorithms for the various datasets in \autoref{fig:datasets}.

\begin{table*}[ht]
\centering
\setlength{\tabcolsep}{20pt}
\renewcommand{\arraystretch}{1.3}
\caption{Comparison of algorithm performance on various datasets}
\begin{tabular}{|l|l|c|}
\hline
Dataset & Algorithm & MAE \\ \hline
\multirow{5}{*}{Auto} & Random Forest & 2.048 \\
& Decision Tree & 2.684 \\
& \textit{k}-NN & 3.271 \\
& XG Boost & 2.141 \\
& Proposed with $\kappa_1=11, \kappa_2=6$ & \textbf{1.912} \\ \hline
\multirow{5}{*}{Student Performance} & Random Forest & 1.529 \\
& Decision Tree & 1.717 \\
& \textit{k}-NN & 1.868 \\
& XG Boost & 1.804 \\
& Proposed with $\kappa_1=8, \kappa_2=8$ & \textbf{1.045} \\ \hline
\multirow{5}{*}{Energy Y2} & Random Forest & 1.501 \\
& Decision Tree & 1.678 \\
& \textit{k}-NN & 2.055 \\
& XG Boost & 1.275 \\
& Proposed with $\kappa_1=12, \kappa_2=5$ & \textbf{1.118} \\ \hline
\multirow{5}{*}{Energy Y1} & Random Forest & 0.587 \\
& Decision Tree & 0.545 \\
& \textit{k}-NN & 1.523 \\
& XG Boost & 0.461 \\
& Proposed with $\kappa_1=2, \kappa_2=24$ & \textbf{0.347}  \\ \hline
\multirow{5}{*}{Iris} & Random Forest & 0.356 \\
& Decision Tree & 0.413 \\
& \textit{k}-NN & 0.348 \\
& XG Boost & 0.416 \\
& Proposed with $\kappa_1=7, \kappa_2=50$ & \textbf{0.269} \\ \hline
\multirow{5}{*}{Concrete} & Random Forest & 5.009 \\
& Decision Tree & 6.944 \\
& \textit{k}-NN & 7.970 \\
& XG Boost & 4.818 \\
& Proposed with $\kappa_1=7, \kappa_2=10$ & \textbf{4.382} \\ \hline
\multirow{5}{*}{Wine Quality} & Random Forest & 0.586 \\
& Decision Tree & 0.755 \\
& \textit{k}-NN & 0.726 \\
& XG Boost & 0.655 \\
& Proposed with $\kappa_1=0.5, \kappa_2=13$ &  \textbf{0.584} \\ \hline
\multirow{5}{*}{Average} & Random Forest & 1.659 \\
& Decision Tree & 2.105 \\
& \textit{k}-NN & 2.537 \\
& XG Boost & 1.653 \\
& Proposed & \textbf{1.380} \\ \hline
\end{tabular}
\label{tab:dataset_results}
\end{table*}

\subsection{Run-Time Analysis}

The proposed model performs well against all other models used in this comparison but it requires more computational time.  We did some run time testing for the datasets for the various algorithms using {\tt time.perf\_counter()} which is a function in the `time' module of Python's standard library. It is used to measure the time a block of code typically takes to run. The function returns the value, in fractional seconds, of a high-resolution timer. On average, the proposed approach took approximately 1.56 times longer than XG Boost and about 274 times longer than the other algorithms. However, we can perform computation optimizations that will reduce the run-time significantly. We plan to explore ways to more efficiently determine the optimal $\kappa$ values.

\subsection{Parameter Optimization}

\begin{figure*}[ht]
  \includegraphics[width=\textwidth]{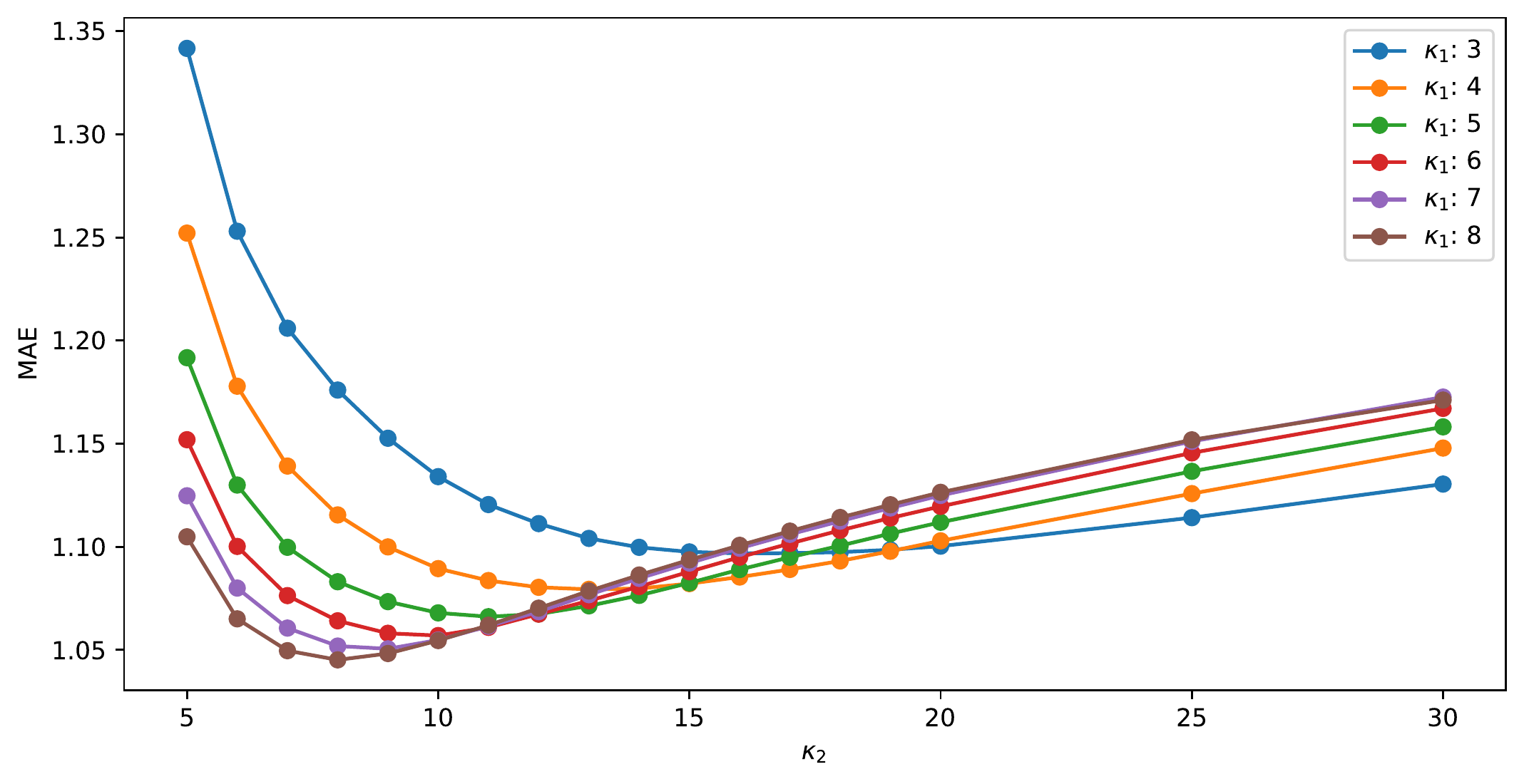}
  \caption{MAE vs $\kappa$ values for the Student Dataset}
  \label{fig:studentkappagraph}
\end{figure*}

According to \cite{Hosein2022}, the error appears to be a convex function of $\kappa$. As $\kappa$ increases, the error decreases up to a certain point and then the error increases after this point. Therefore, an optimal value can be found that minimizes the error. Our approach uses two different $\kappa$ values, one for imputing the values and one for predicting. This allows for more flexibility in the model since each value serves a different purpose and allows them to optimize both aspects independently to minimize the error. In some cases, the optimal $\kappa$ value for the first part may not be best for the predicting part and vice versa. In \autoref{fig:studentkappagraph}, we can clearly see the pattern of the error. The MAE decreases to a minimum and then increases after $\kappa_2$ = 8. In our approach, we used a single value for all the features when imputing. However, we note that the parameter can be further optimised by using different values for different features. This further optimisation will lead to an even more complex model but may yield better results so we intend to explore this in the future.

\section{\uppercase{Discussion}}

We compared the Random Forest, Decision Tree, \textit{k}-Nearest Neighbours (\textit{k}-NN), XG Boost and the proposed algorithm on seven diverse datasets from UCI. The datasets were from various fields of study and consist of a combination of categorical and ordinal features. Our proposed approach uses two hyper-parameters to optimize predictions. The average mean absolute error of the proposed approach is 45.6 \% lower than \textit{k}-NN, 34.4\% lower than Decision Tree, 16.8\% lower than the Random Forest and 16.5\% lower than XG Boost. The proposed approach achieves the lowest MAE for all datasets. These results illustrate the value and potential of the proposed approach.

\section{\uppercase{Conclusions and Future Work}}
\label{sec:conclusion}

We present a robust approach that can be used for any regression problem. The approach is based on a weighted average of the target values of the training points where the weights are determined by the inverse of the Euclidean distance between the test point and the training points raised to the power of a parameter $\kappa$.
As shown in Figure \ref{fig:datasets} the proposed algorithm surpasses the traditional algorithms in each dataset. Its performance indicates that the proposed method is a promising approach for solving regression tasks and should be considered as a strong candidate for future applications. However, there is significant room for improvement of this algorithm. Future work can include using different $\kappa$ values for each feature and exploring heuristic methods to determine these values which may result in even better performance. Also, since the algorithm's computations can be done in parallel (i.e., the grid search over the $\kappa$ space), the run time can also be considerably decreased. 

\bibliographystyle{apalike}
{\small
\bibliography{data2023}}

\begin{thebibliography}{}

\bibitem[Banerjee, 2020]{Banerjee2020}
Banerjee, P. (2020).
\newblock Comprehensive guide on feature selection.

\bibitem[Breiman, 2001]{Breiman2001}
Breiman, L. (2001).
\newblock Random forests.
\newblock {\em Machine learning}, 45(1):5--32.

\bibitem[Brownlee, 2021]{Brownlee2021}
Brownlee, J. (2021).
\newblock Recursive feature elimination (rfe) for feature selection in python.

\bibitem[Chen and Guestrin, 2016]{Chen2016}
Chen, T. and Guestrin, C. (2016).
\newblock Xgboost: A scalable tree boosting system.
\newblock In {\em Proceedings of the 22nd ACM SIGKDD International Conference
  on Knowledge Discovery and Data Mining}, pages 785--794. ACM.

\bibitem[Cortez, 2014]{Cortez2014}
Cortez, P. (2014).
\newblock {Student Performance}.
\newblock UCI Machine Learning Repository.
\newblock {DOI}: \url{10.24432/C5TG7T}.

\bibitem[Cortez et~al., 2009]{Cortez2009}
Cortez, P., Cerdeira, A., Almeida, F., Matos, T., and Reis, J. (2009).
\newblock {Wine Quality}.
\newblock UCI Machine Learning Repository.
\newblock {DOI}: \url{10.24432/C56S3T}.

\bibitem[Fisher, 1988]{Fisher1988}
Fisher, R. A. \&~Fisher, R. (1988).
\newblock {Iris}.
\newblock UCI Machine Learning Repository.
\newblock {DOI}: \url{10.24432/C56C76}.

\bibitem[Gooljar, 2023]{Gooljar2023}
Gooljar, S. (2023).
\newblock Comparison of the performance of a proposed algorithm to other
  algorithms.
\newblock GitHub repository.

\bibitem[Hosein, 2022]{Hosein2022}
Hosein, P. (2022).
\newblock A data science approach to risk assessment for automobile insurance
  policies.
\newblock {\em International Journal of Data Science and Analytics}, pages
  1--12.

\bibitem[Quinlan, 1986]{Quinlan1986}
Quinlan, J.~R. (1986).
\newblock Induction of decision trees.
\newblock {\em Machine learning}, 1(1):81--106.

\bibitem[Quinlan, 1993]{Quinlan1993}
Quinlan, R. (1993).
\newblock {Auto MPG}.
\newblock UCI Machine Learning Repository.
\newblock {DOI}: \url{10.24432/C5859H}.

\bibitem[Tsanas, 2012]{Tsanas2012}
Tsanas, Athanasios \&~Xifara, A. (2012).
\newblock {Energy efficiency}.
\newblock UCI Machine Learning Repository.
\newblock {DOI}: \url{10.24432/C51307}.

\bibitem[Yeh, 2007]{Yeh2007}
Yeh, I.-C. (2007).
\newblock {Concrete Compressive Strength}.
\newblock UCI Machine Learning Repository.
\newblock {DOI}: \url{10.24432/C5PK67}.

\end{thebibliography}
\end{document}